\pdfoutput=1
\documentclass[11pt]{article}

\usepackage[preprint]{acl}

\usepackage{times}
\usepackage{latexsym}

\usepackage[T1]{fontenc}
\usepackage[utf8]{inputenc}

\usepackage{microtype}

\usepackage{inconsolata}

\usepackage{graphicx}

\usepackage{multirow}
\usepackage{amsmath, bm, amssymb}
\usepackage{tabularx}
\usepackage{makecell} % 用于换行列名
\usepackage{booktabs} % 用于更美观的表格线
\usepackage{hyperref}

\newcommand{\mysection}[1]{\S\,\ref{#1}}

\title{Decoding Knowledge Attribution in Mixture-of-Experts: A Framework of Basic-Refinement Collaboration and Efficiency Analysis}

\author{Junzhuo Li\textsuperscript{1}\textsuperscript{2}%\footnotemark[2]
, Bo Wang\textsuperscript{1}, Xiuze Zhou\textsuperscript{1}\textsuperscript{2}, Peijie Jiang\textsuperscript{3}, Jia Liu\textsuperscript{3}  \and Xuming Hu\textsuperscript{1}\textsuperscript{2}\footnotemark[1]\\
        \textsuperscript{1}The Hong Kong University of Science and Technology (Guangzhou)\\%, Tianjin, China \\
        \textsuperscript{2} The Hong Kong University of Science and Technology \\
        \textsuperscript{3}Ant Group \\
        \texttt{\{jz.li, bo.wang, xz.zhou\}@connect.hkust-gz.edu.cn} \\
        \texttt{xuminghu@hkust-gz.edu.cn}
        }

\begin{document}
\maketitle
\begin{abstract}
The interpretability of Mixture-of-Experts (MoE) models, especially those with heterogeneous designs, remains underexplored. Existing attribution methods for dense models fail to capture dynamic routing-expert interactions in sparse MoE architectures. To address this issue, we propose a cross-level attribution algorithm to analyze sparse MoE architectures (Qwen 1.5-MoE, OLMoE, Mixtral-8x7B) against dense models (Qwen 1.5-7B, Llama-7B, Mistral-7B). Results show MoE models achieve 31\% higher per-layer efficiency via a ``mid-activation, late-amplification'' pattern: early layers screen experts, while late layers refine knowledge collaboratively. Ablation studies reveal a ``basic-refinement'' framework—shared experts handle general tasks (entity recognition), while routed experts specialize in domain-specific processing (geographic attributes). Semantic-driven routing is evidenced by strong correlations between attention heads and experts ($r=0.68$), enabling task-aware coordination. Notably, architectural depth dictates robustness: deep Qwen 1.5-MoE mitigates expert failures (e.g., 43\% MRR drop in geographic tasks when blocking top-10 experts) 
through shared expert redundancy, whereas shallow OLMoE suffers severe degradation (76\% drop). 
Task sensitivity further guides design: core-sensitive tasks (geography) require concentrated expertise, while distributed-tolerant tasks (object attributes) leverage broader participation. These insights advance MoE interpretability, offering principles to balance efficiency, specialization, and robustness.
\end{abstract}
\renewcommand{\thefootnote}{\fnsymbol{footnote}}
% \footnotetext[]{Correspongding authors.}
% \footnotetext[2]{Work done while this author was an intern at Ant Group.}
\footnotetext[1]{Corresponding author.}
% \section{Introduction}

% These instructions are for authors submitting papers to *ACL conferences using \LaTeX. They are not self-contained. All authors must follow the general instructions for *ACL proceedings,\footnote{\url{http://acl-org.github.io/ACLPUB/formatting.html}} and this document contains additional instructions for the \LaTeX{} style files.

% The templates include the \LaTeX{} source of this document (\texttt{acl\_latex.tex}),
% the \LaTeX{} style file used to format it (\texttt{acl.sty}),
% an ACL bibliography style (\texttt{acl\_natbib.bst}),
% an example bibliography (\texttt{custom.bib}),
% and the bibliography for the ACL Anthology (\texttt{anthology.bib}).

\section{Introduction}

In recent years, transformer-based large language models (LLMs) \citep{vaswani2017attention, brown-2020-language, touvron2023llama, qwen2023dense} have significantly enhanced task performance by scaling up parameters, yet at the cost of prohibitive computational demands. To balance efficiency and performance, Mixture-of-Experts (MoE) architectures \citep{fedus-2021-switch, lepikhin2021gshard, shen2023mixture, albert-2024-mixtral, dai-etal-2024-deepseekmoe, sun2024hunyuanlargeopensourcemoemodel} activate subsets of experts per input, reducing overhead while maintaining capability. While existing research has prioritized optimizing routing mechanisms and expert module efficiency (e.g., top-$k$ gating \citep{fedus-2021-switch, lepikhin2021gshard}), a critical question remains unanswered: \textit{How do experts collaboratively process and refine knowledge within MoE models?}

This gap in interpretability is particularly pronounced in models with heterogeneous designs, such as shared expert modules \citep{dai-etal-2024-deepseekmoe}. Despite their widespread adoption \citep{deepseekai2024deepseekv3technicalreport, qwen_moe}, the roles of shared experts—whether they serve as universal feature extractors or redundant backups—lack empirical validation. The opacity of expert collaboration not only hampers credibility assessment (e.g., verifying if experts specialize in specific knowledge domains) but also forces model designers to rely on trial-and-error strategies, limiting systematic improvements. For instance, fundamental issues like the distribution of shared experts across layers and their impact on knowledge consistency remain unexplored.

Two key challenges compound this problem: (1) Existing attribution methods \citep{geva-etal-2021-transformer, dai-etal-2022-knowledge, geva-etal-2022-transformer, wu-etal-2023-depn} are designed for dense models and fail to capture dynamic interactions between routed experts and shared modules.
(2) Comparative studies on heterogeneous MoE architectures are scarce, leaving functional assumptions (e.g., ``shared experts handle general tasks'') unverified.

\begin{figure}[t]
  \includegraphics[width=\columnwidth]{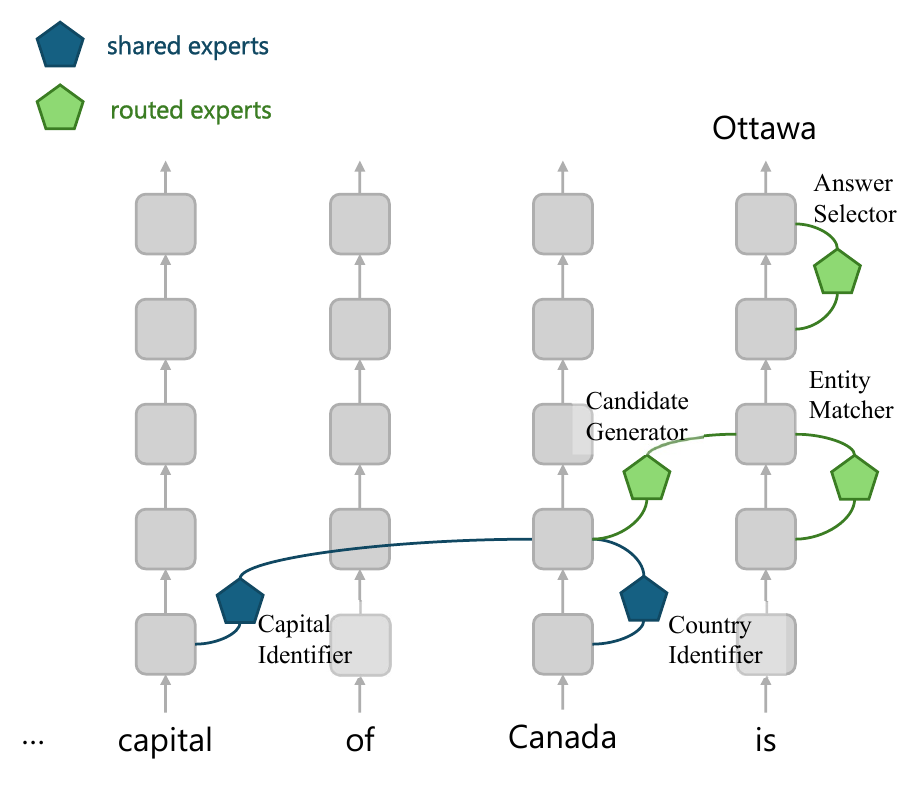}
  \caption{Schematic of the MoE expert layer: Shared experts perform foundational tasks (e.g., the Capital Identifier recognizes “capital,” the Country Identifier recognizes “Canada”), while routed experts execute refinement steps—generating candidate capitals, matching them to “Canada,” and selecting the final answer (“Ottawa”).}
  \label{fig:expert_role}
\end{figure}

In this paper, we address these gaps by proposing a cross-level attribution algorithm for MoE models, extending dense model methods to analyze both macro architectures (Mixtral-8x7B \citep{albert-2024-mixtral}, Qwen 1.5-MoE \citep{qwen_moe}, OLMoE \cite{muennighoff2024olmoe}) and micro behaviors. Our experiments uncover four pivotal insights. First, MoE models significantly outperform dense counterparts in per-layer efficiency—Qwen 1.5-MoE achieves a layer efficiency of 0.294 with 24 layers, surpassing Qwen 1.5-7B (0.213 with 32 layers), a 37\% improvement attributed to concentrated knowledge processing in early layers. Second, we identify a hierarchical "basic-refinement" collaboration: shared experts act as generalists for foundational tasks like entity localization, while routed experts specialize in refining domain-specific attributes (e.g., linking "Canada" to "Ottawa"). Third, semantic-driven routing is validated through strong temporal correlations between attention heads and expert selection ($r=0.68, p<0.0014$), indicating that attention mechanisms guide experts toward task-relevant features. Finally, robustness varies starkly by architecture: blocking top-10 experts in deep Qwen 1.5-MoE causes a 43\% MRR drop in geographic tasks, whereas shallow OLMoE suffers a 76\% decline, underscoring the critical role of depth and shared experts in redundancy design.

\paragraph{Contributions} Our main contributions are as follows:
\begin{itemize}
\item \textbf{A cross-level attribution algorithm for MoE models}, bridging the gap in interpreting sparse architectures by dynamically weighting expert and attention contributions.
\item \textbf{Empirical validation of MoE efficiency and collaboration patterns}, including layer efficiency gains, a “mid-activation, late-amplification” processing paradigm, and semantic-driven expert routing.
\item \textbf{Design implications for robust MoE architectures}, demonstrating that deep models with shared experts balance specialization and redundancy, particularly critical for complex tasks like geographic reasoning.
\end{itemize}

\begin{table*}[t]
\centering
\begin{tabularx}{\textwidth}{lcccccc}
\toprule
\makecell[l]{Model} & \makecell[c]{HIT@10} & \makecell[c]{MRR} & \makecell[c]{Total \\ FFN Gain} & \makecell[c]{Total \\ Attn Gain} & \makecell[c]{Peak Gain \\ Layer Position \\ (Relative Position\%)} & \makecell[c]{Layer \\ Efficiency \\ (FFN Gain / \\ Num. of Layers)} \\
\midrule
Llama-7B & 0.90 & 0.70 & 5.16 & 4.03 & 24.82 (77.6\%) & 0.161 \\
Qwen 1.5-7B & 0.79 & 0.62 & 6.49 & 4.14 & 27.08 (90.2\%) & 0.203 \\
Mistral-7B & 0.88 & 0.71 & 6.74 & 2.96 & 26.64 (83.2\%) & 0.211 \\
Qwen 1.5-MoE & 0.85 & 0.63 & 7.36 & 3.18 & 20.36 (84.8\%) & 0.307 \\
OLMoE & 0.83 & 0.64 & 4.98 & 4.40 & 13.53 (84.6\%) & 0.311 \\
Mixtral-8x7B & 0.90 & 0.73 & 6.79 & 3.03 & 26.66(83.3\%) & 0.212 \\

\bottomrule
\end{tabularx}
\caption{Comparison of model architectures on performance metrics, component contributions, and layer efficiency across dense and MoE models.}
\label{tab:comparison}
\end{table*}

\section{Methodology}
In this section, we first describe the MoE model architecture, followed by the knowledge attribution method. %, and finally, we outline the experimental setup.
More details for model settings and experimental settings (i.e., dataset, evaluation metrics) can be found in Appendix~\ref{sec:appendix_experiment}.

\subsection{MoE Architecture}
In the decoder-only MoE architecture, each Transformer decoder block replaces the conventional Feed-Forward Network (FFN) with an MoE layer, enhancing model capacity via dynamic routing. Given an input sequence \( X = (x_1, x_2, \dots, x_T) \), the tokens are embedded to produce the initial hidden state \( \bm{h}^0 = \text{Embed}(X) \). We omit bias and layer normalization \citep{ba-2016-layer} following \citet{geva-etal-2022-transformer}.

Each decoder layer consists of a self-attention mechanism and an MoE layer. At layer \( l \), the attention output for token \( x_i \) is computed as:
\begin{equation}
    \bm{A}_i^l = \sum_{j=1}^{H} \text{ATTN}_j^l(\bm{h}_1^{l-1}, \bm{h}_2^{l-1}, \dots, \bm{h}_T^{l-1}),
\end{equation}
where \( H \) is the number of attention heads. The intermediate representation \( \bm{u}_i^l = \bm{A}_i^l + \bm{h}_i^{l-1} \) is routed to \( N \) experts via a gating network, which computes the routing probabilities:
\begin{equation}\label{req:routing}
    \bm{g}_i^l = \text{softmax}(\bm{W}_g^l \bm{u}_i^l + \bm{b}_g^l),
\end{equation}
where \( \bm{W}_g^l \) and \( \bm{b}_g^l \) are learnable parameters. Each expert \( \mathcal{E}_j^l \) is a feed-forward network:
\begin{equation}
    \mathcal{E}_j^l(\bm{u}_i^l) = \bm{W}_{j,2}^l \phi(\bm{W}_{j,1}^l \bm{u}_i^l + \bm{b}_j^l),
\end{equation}
where \( \phi(\cdot) \) is a nonlinear activation function \cite{relu, leaky_relu, gelu}. The final output for token \( x_i \) at layer \( l \) is computed by a weighted sum of the top \( k \) experts' outputs:
\begin{equation}
    \bm{F}_i^l = \sum_{j=1}^{k} g_{i, j}^l \cdot \mathcal{E}_j^l(\bm{u}_i^l).
\end{equation}
Some MoEs include shared experts that are always selected \citep{dai-etal-2024-deepseekmoe}, which results in:
\begin{equation}
\bm{F}_i^l = \sum_{j=1}^{k} g_{i,j}^l \cdot \mathcal{E}_j^l(\bm{u}_i^l) + g_{i, s}^l \cdot \mathcal{E}_s^l(\bm{u}_i^l).
\end{equation}
Finally, the output at position \( i \) of the \( l \)-th layer is:
\begin{equation}
    \bm{h}_i^l = \bm{h}_i^{l-1} + \bm{A}_i^l + \bm{F}_i^l.
\end{equation}
% After the output \( \bm{h}_i^L \) is computed in the final layer, it is passed through the decoder's output layer. 
The final hidden states \( \bm{h}_i^L \) need to go through the unembedding process to map the high-dimensional representations back to the vocabulary space. Specifically, for each position \( i \), the final output token vector \( \bm{y}_i \) is obtained by multiplying the hidden state \( \bm{h}_i^L \) with the unembedding matrix \( \bm{W}_{\text{unembed}} \) and passing it through a softmax function.

\subsection{Neuron-Level Attribution}
We adopt the neuron-level attribution methods from \citet{yu-ananiadou-2024-neuron} for the FFN and attention layers and extend them for MoE models.  For attention neurons \( \bm{v}^l_A \), a vector in $l$-th Attention layer, the importance score is:
\begin{equation}
    \mathcal{I}(\bm{v}^l_A) = \log p(x_i | \bm{v}^l_A + \bm{h}^{l-1}) - \log p(x_i | \bm{h}^{l-1}),
\end{equation}
% where \( \bm{v}^l_A \) is a vector (neuron) in $l$-th Attention layer.

For a token $x_i$ The importance score for FFN neurons is defined as the log probability increase:
\begin{equation}
    \mathcal{I}(\bm{v}^l_F) = \log p(x_i | \bm{v}^l_F+ \bm{u}^{l}) - \log p(x_i | \bm{u}^{l}),
\end{equation}
where \( \bm{v}^l_F \) is a vector (neuron) in $l$-th FFN layer. %is the output of the FFN at layer \( l \).% and \( h^{l-1} \) is the previous hidden state.

In MoE models, we extend these methods to account for dynamic routing and expert contributions. The importance score for MoE expert neurons is computed as:
\begin{equation}
    \mathcal{I}(\bm{v}^l_{\mathcal{E}_j}) = \log p(x_i | g_{i,j}^l \bm{v}^l_{\mathcal{E}_j} + \bm{u}^{l}) - \log p(x_i | \bm{u}^{l}),
\end{equation}
where \( g_{i,j}^l \) is the gating probability for expert \( \mathcal{E}_j \), and \( v^l_{\mathcal{E}_j} \) is the neuron from expert \( \mathcal{E}_j \) of $l$-th layer.

Through these modifications, we are able to accurately attribute importance not only to the conventional FFN and attention neurons but also to the individual experts and their specific contributions within the MoE framework. This extension allows us to capture the complex interactions between experts and tokens, providing deeper insights into the behavior of MoE models and how they utilize expert knowledge.

The importance score represents the influence of a neuron on the final prediction, reflecting the extent of its contribution during the model’s inference process. More specifically, it can be regarded as the gain introduced by the module to the overall model performance, revealing the role of different neurons or experts in the model’s output. This metric aids in understanding how the model utilizes the knowledge of various components and the significance of each module for specific tasks.

\section{Comparative Analysis: Efficiency and Knowledge Processing in MoE vs. Dense Models}

In this section, we compare three MoE architectures (Qwen 1.5-MoE, Mixtral, and OLMoE) with two dense models (Qwen 1.5-7B, Llama-7B, and Mistral) across several key metrics. Our analysis highlights significant differences in parameter efficiency, knowledge localization, and FFN gain evolution—differences that underscore the unique interpretability advantages of MoE models.

\subsection{Layer Efficiency and Knowledge Concentration in MoE Models}
Table~\ref{tab:comparison} summarizes the performance of the four models, revealing that MoE architectures achieve notably higher per-layer gains. For example, Qwen 1.5-MoE attains a layer efficiency of 0.307 using only 24 layers, whereas its dense counterpart, Qwen 1.5-7B, requires 32 layers to reach an efficiency of just 0.203. Despite a slightly lower overall FFN gain, OLMoE—by virtue of its shallower design—achieves the highest layer efficiency at 0.311, demonstrating a marked optimization in computational density.

The data further indicate that MoE models concentrate their knowledge processing at earlier stages. Specifically, Qwen 1.5-MoE and OLMoE reach their peak FFN gains at 84.8\% and 84.6\% of the network depth, respectively. This “early routing, late refinement” pattern enables these models to perform initial expert filtering in mid-layers and then focus on intensive knowledge refinement in later layers. In contrast, dense models such as Qwen 1.5-7B and Llama-7B display a more gradual accumulation of gains—peaking at 90.2\% and 77.6\% of the network depth—which, while beneficial for cross-layer integration, tends to blur the functional distinctions among layers and reduce interpretability.
\begin{figure}[t]
  \includegraphics[width=\columnwidth]{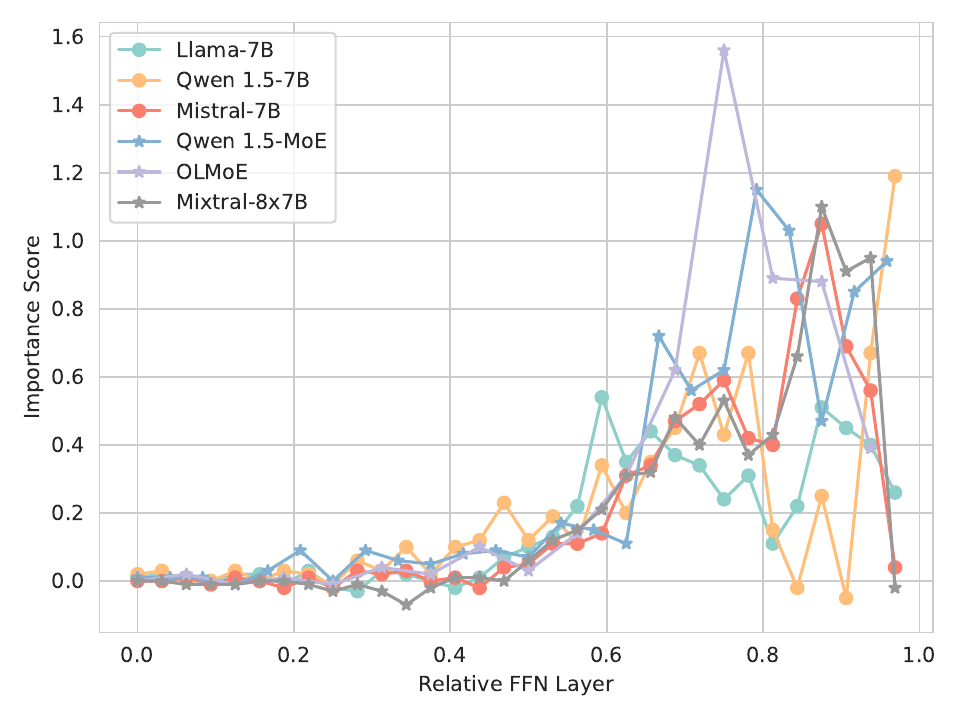}
  \caption{Comparison of layer importance scores distribution in FFN layers.}
  \label{fig:layer_importance_distribution}
\end{figure}

Furthermore, a comparison of attention versus FFN gains underscores divergent processing strategies. Qwen 1.5-MoE, for instance, exhibits a pronounced reliance on specialized expert networks, with an FFN gain of 7.36 contrasted against an attention gain of only 3.18. Dense models, by comparison, distribute gains more evenly (with attention gains of 4.14 and 4.03 for Qwen 1.5-7B and Llama-7B, respectively), complicating the attribution of specific processing roles to individual layers. Notably, OLMoE achieves nearly identical attention and FFN gains (4.98 and 4.40, respectively), reflecting a balanced strategy that integrates both attention mechanisms and expert routing.

% \subsection{FFN Gain Evolution Across Layers}
\subsection{FFN Gain Evolution: Mid-Stage Activation and Late-Stage Amplification}
Figure~\ref{fig:layer_importance_distribution} illustrates the evolution of FFN gains across layers in Llama-7B, Qwen 1.5-MoE, Qwen 1.5-7B , OLMoE, Mistral, and Mixtral. The analysis reveals distinct processing patterns shaped by architectural designs:

Qwen 1.5-MoE exhibits a three-phase dynamic: (1) Early stage (Layers 1–13): Parallel initialization of experts extracts basic features (e.g., entity localization), contributing 6.1\% of the total gain. (2) Mid-stage (Layers 14–19): Dynamic routing activates expert screening, where the top-4 gating mechanism selects task-relevant specialists. This phase contributes 43.5\% of the total gain while minimizing redundant computations. (3) Late stage (Layers 20–24): Collaborative refinement between shared experts (e.g., broad knowledge integration) and routed experts (e.g., attribute association) amplifies gains, accounting for 50.4\% of the total.

In contrast, Qwen 1.5-7B relies on linear accumulation across deeper layers (25–32), gradually integrating knowledge without explicit stage specialization. OLMoE, as a shallow architecture, rapidly peaks at layer 12 (mid-stage) but lacks sustained late-stage refinement due to limited depth, leading to a gradual performance decay.

Notably, Mistral diverges significantly from typical dense models: While other dense architectures exhibit diminished mid-to-late-stage gains, Mistral sustains high FFN activation through later layers, achieving gain levels comparable to MoE models. Mixtral shows strong similarity to Mistral, with both models peaking sharply at layer 28 before experiencing rapid performance decay. This parallel trajectory—absent in other architectures—suggests potential initialization-driven dynamics influencing their gain structures.  

This structured processing enables MoE models to achieve higher per-layer efficiency. For example, Qwen 1.5-MoE’s late-stage layers (20–24) generate a gain of 1.45 per layer, compared to Qwen 1.5-7B’s 0.68 per layer in layers 25–32. The efficiency stems from task-specialized routing and cross-expert collaboration, contrasting with dense models’ uniform parameter utilization. The exceptional late-stage performance of Mistral and its resonance with Mixtral, however, indicate that initialization strategies \citep{lo-etal-2025-closer} and architectural synergies may further modulate gain dynamics beyond conventional dense/MoE dichotomies.

\begin{figure}[t]
  \includegraphics[width=\columnwidth]{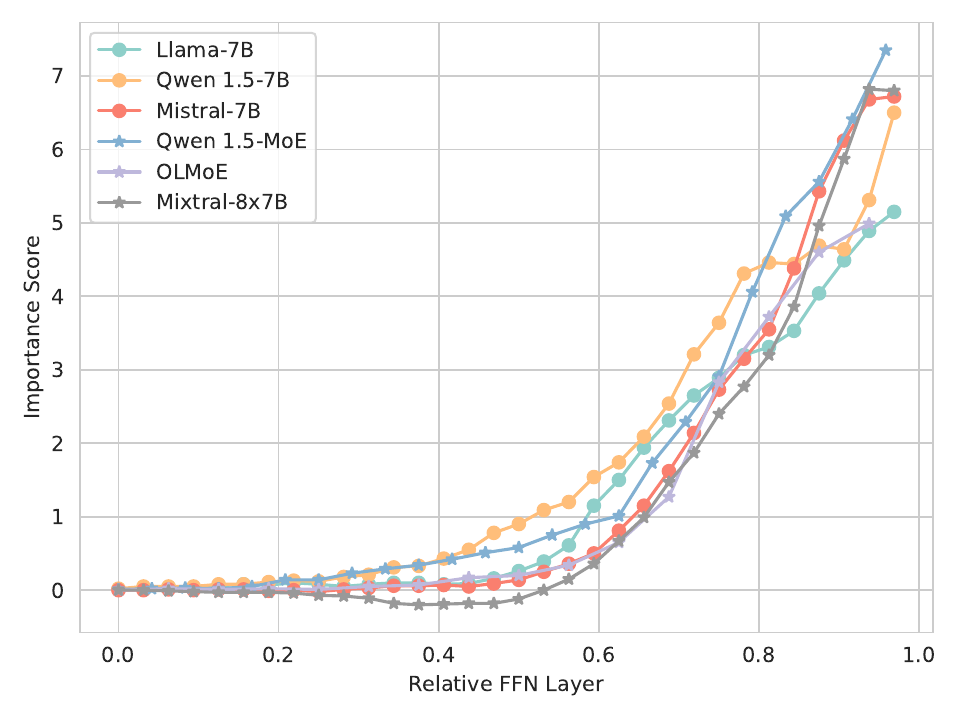}
  \caption{Cumulative FFN layers contribution curves.}
  \label{fig:cumu_contribution}
\end{figure}

% \subsection{Stage-wise Contribution Analysis}
\subsection{Cross-Model Comparison of Stage Contributions}

Having established the stage-wise processing patterns, we next compare their contributions across architectures.
To quantify architectural differences, Figure~\ref{fig:cumu_contribution} and Table~\ref{tab:comparison} compare cumulative contributions across early, mid-, and late stages. 

\paragraph{Mid-stage dominance in MoE}  Qwen 1.5-MoE achieves 41.7\% mid-stage contribution (absolute gain: 2.93), surpassing Qwen 1.5-7B’s 37.5\% (2.56). OLMoE’s shallow design further amplifies mid-stage reliance (52.2\% contribution), but limited depth restricts total gain (4.79 vs. Qwen 1.5-MoE’s 7.05).

\paragraph{Late-stage amplification in deep MoE} Qwen 1.5-MoE’s 24-layer architecture enables extended refinement, with 50.7\% late-stage gain (3.57) versus Qwen 1.5-7B’s 47.6\% (3.25). OLMoE’s 16-layer design results in rapid late-stage decay (12.5\% contribution post-layer 12), highlighting the trade-off between depth and efficiency.

\paragraph{Architectural impact on processing strategies} Deep MoE models (e.g., Qwen 1.5-MoE) prioritize collaborative refinement in late layers, while shallow MoE architectures (e.g., OLMoE) focus on rapid mid-stage decisions. Dense models’ dispersed gains hinder task specialization—Qwen 1.5-7B requires 32 layers to achieve comparable total gains, indicating inefficiency in cross-layer integration.

These results underscore that MoE’s efficiency arises from stage-specialized processing, yet its effectiveness depends on balancing depth and expert collaboration. A detailed analysis of robustness implications (e.g., expert blocking effects) is provided in \mysection{sec:block_expert}.%Section 5.2.

\section{Expert Collaboration: Basic-Refinement via Semantic Routing}

This section investigates the collaborative mechanisms between shared and routed experts in MoE models. Through systematic ablation experiments, we validate the necessity of expert collaboration for robust knowledge processing (\mysection{sec:expert_role}) and reveal how semantic-driven routing coordinates attention heads and experts into a hierarchical framework (\mysection{sec:attn_expert}). 

\subsection{Expert Role Specialization}\label{sec:expert_role}

\begin{table}[t]
  \centering
  \begin{tabular}{l c c}
    \toprule
    Model Config & HIT@10  & MRR \\
    \midrule
    Only Top1  & 0 & 0 \\ % (w/o shared expert + top 1 routed expert)
    Only Top2 & 0 & 0 \\
    Only Shared  & 0.03 & 0.01 \\ %(shared expert + top 0 routed expert)
    Shared + 2nd  & 0.04 & 0.01\\ % (shared expert + only 2nd expert activated)
    Shared + Top1  & 0.82 & 0.59\\ %(w/ shared expert + top 1 routed expert)
    Shared + Top2  & 0.83 & 0.63\\ %(w/ shared expert + top 1 routed expert)
    Shared + Top4 (default) & 0.85 & 0.63\\
    \bottomrule
  \end{tabular}
  \caption{Experimental results for different Qwen 1.5-MoE configurations, highlighting the effects of the shared expert and top-1 routing.}
  \label{tab:qwen_result}
\end{table}

% mixtral top0, top1, top2(default)
\begin{table}[t]
  \centering
  \begin{tabular}{l c c}
    \toprule
    Model Config & HIT@10  & MRR \\
    \midrule
    Top0  & 0.00 & 0.00\\ 
    Top1  & 0.86 & 0.67\\ 
    Top2 (default) & 0.90 & 0.73\\
    \bottomrule
  \end{tabular}
  \caption{Experimental results for different Mixtral configurations.}
  \label{tab:mixtral_result}
\end{table}

\begin{figure}[t]
  \includegraphics[width=\columnwidth]{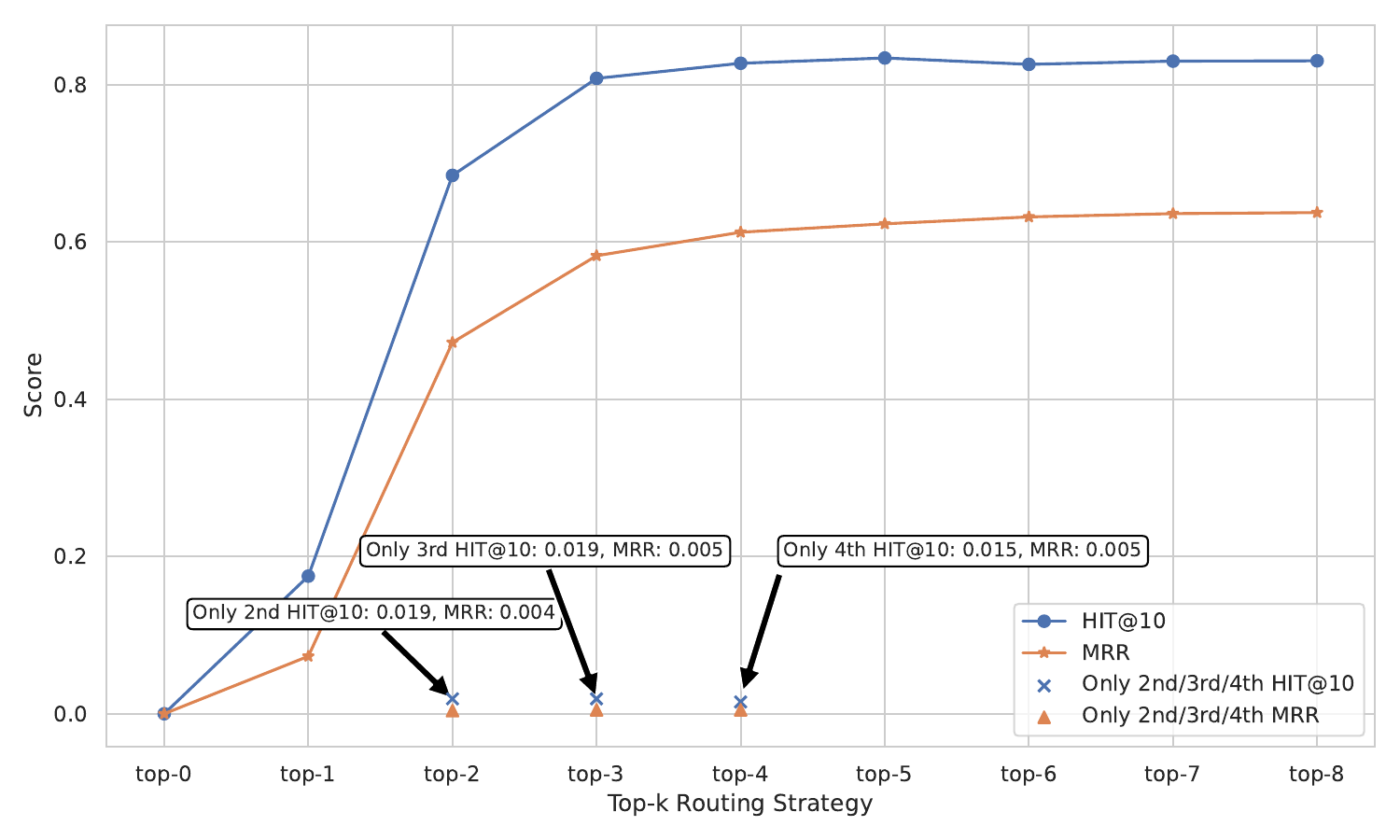}
  \caption{OLMoE’s performance on knowledge prediction tasks when using different routing strategies.}
  \label{fig:olmoe_ablation_expert}
\end{figure}

Our ablation studies demonstrate that effective collaboration between shared and routed experts is fundamental to MoE performance. In both Qwen 1.5-MoE (Table~\ref{tab:qwen_result})—which combines a shared expert with routed specialists—and OLMoE (Figure~\ref{fig:olmoe_ablation_expert}) and Mixtral (Table~\ref{tab:mixtral_result})—which relies solely on routed experts—isolating individual experts severely degrades model capability. For example, when only a single expert is activated, OLMoE’s HIT@10 drops to a range of 0.15–0.19 with near-zero MRR, while Qwen 1.5-MoE achieves a mere HIT@10 of 0.03 when restricted to its shared experts and completely fails when limited to the top routed expert. These results underscore that complex knowledge tasks cannot be managed by isolated experts; instead, they require synergistic interactions between generalists and specialists.

A critical observation is the diminishing returns of increasing expert participation. In OLMoE, activating three experts elevates HIT@10 from 0.18 (single expert) to 0.81, but adding a fourth expert yields only a marginal gain (0.81 to 0.83). Similarly, Qwen 1.5-MoE sees limited improvement when expanding beyond its shared experts and top routed expert—HIT@10 rises from 0.82 to 0.85 and MRR from 0.59 to 0.63. This suggests the existence of an “effective expert threshold”: core knowledge is predominantly captured by a small subset of experts, with additional modules contributing minimally.

Architectural differences further amplify these patterns. Qwen 1.5-MoE’s shared experts handles cross-domain fundamentals—such as entity recognition and syntactic parsing—freeing routed experts to specialize in domain-specific refinement. This division allows Qwen 1.5-MoE to achieve high performance with minimal experts while maintaining fault tolerance; blocking a routed expert reduces HIT@10 by only 3\% (0.85 to 0.82). In contrast, OLMoE’s lack of a shared expert forces it to activate multiple routed experts (typically three) for comparable accuracy, making it vulnerable to routing failures. For instance, blocking a critical expert in OLMoE degrades MRR by over 30\%, as shown in Figure~\ref{fig:olmoe_ablation_expert}.

These findings collectively support a “basic-refinement” framework for MoE design (as shown in Figure~\ref{fig:expert_role}): (1) Shared experts act as generalists, providing robust foundational processing. (2) Routed experts specialize in refining domain-specific knowledge.
This hierarchical collaboration not only enhances efficiency but also improves interpretability, offering practical guidance for optimizing MoE architectures under computational constraints. For instance, prioritizing shared experts in early layers and routed experts in deeper layers could balance performance and resource usage—a strategy validated by Qwen 1.5-MoE’s success.

\subsection{Semantic-Driven Attention Head and Expert Selection Mechanism} \label{sec:attn_expert}

In large-scale MoE models, the interaction between attention heads and experts operates as a tightly coordinated, semantic-driven collaboration. This mechanism ensures that task-specific knowledge is dynamically routed and refined across layers (Figure~\ref{fig:expert_role}). Take Qwen 1.5-MoE as an example, the process begins in the early to mid-layers (layers 18–20), where attention heads such as $\mathcal{H}^{18}_{10}$
focus on encoding foundational semantic features—for example, identifying entities like ``Canada'' as a country. These features are stored and processed by activated experts (e.g., $\mathcal{E}^{18}_{38}$), which serve as repositories of domain-specific knowledge.

As processing progresses to layer 20, the routing mechanism enters a critical phase. Here, attention heads like $\mathcal{H}^{19}_{12}$ align token-level semantics with expert capabilities, selectively activating the top-4 specialists through gating probabilities (Eq.~\ref{req:routing}). For instance, the head $\mathcal{H}^{19}_{12}$  may prioritize expert 
$\mathcal{E}^{20}_{59}$) for tasks involving geographic relations, leveraging its stored knowledge of country attributes. This semantic alignment minimizes redundant computations while maximizing task relevance.

In the deeper layers (21–23), the collaboration shifts toward refinement. Specialized experts such as $\mathcal{E}^{21}_{48}$) work under the guidance of attention heads like $\mathcal{H}^{22}_{13}$) to generate precise outputs—for example, linking ``Canada'' to its capital ``Ottawa''. To validate this hierarchical pipeline, we analyzed temporal correlations between attention heads and experts. The results reveal statistically significant synchronization: the correlation coefficient between $\mathcal{E}^{18}_{10}$ and $\mathcal{E}^{20}_{59}$ reaches $r=0.68\quad(p<0.001)$, while $\mathcal{H}^{23}_{15}$ and $\mathcal{E}^{23}_{20}$ exhibit $r=0.62\quad(p<0.01)$ (Table~\ref{tab:attn_head_to_expert}). These findings confirm that attention heads actively steer expert selection in a semantics-aware manner, contrasting sharply with dense models’ static parameter usage.

\subsection{Causality Analysis}
While the above temporal correlations provide strong evidence of semantic coordination between attention heads and experts, we recognize that correlation alone does not establish a causal relationship. To verify that attention heads actively drive expert selection, we designed and ran three targeted interventions:

\paragraph{Attention Head Suppression}
   We selectively suppressed the activity of head $\mathcal{H}^{19}_{12}$ in Qwen during inference. This intervention reduced the average gating probability for the top-4 experts by 54\% and led to a 29\% drop in mean reciprocal rank (MRR), demonstrating that disabling a single key head disrupts downstream expert routing.

\paragraph{Expert Forcing}
   In a complementary experiment, we forcibly activated expert $\mathcal{E}^{20}_{59}$ regardless of the gating probabilities. This intervention recovered 85\% of the original MRR, indicating that reinstating the correct expert can largely restore model performance even when upstream signals are perturbed.

\paragraph{Path Analysis via Integrated Gradients}
   We applied Integrated Gradients to trace the decision path from attention heads to experts. On average, 28\% of the attribution for expert activations (e.g. the route $\mathcal{H}^{22}_{13} \to \mathcal{E}^{21}_{48}$) could be directly linked back to specific attention heads, quantifying their direct influence in the routing computation.

Together, these causal interventions confirm that attention heads do not merely co-activate with experts but in fact guide their selection in a semantics-aware, hierarchical pipeline.

\begin{table}[t]
  \centering
  \resizebox{0.48\textwidth}{!}{
  \begin{tabular}{l c c}
    \toprule
    Relation & \makecell[c]{High-Imp. \\ Attention Head}   & \makecell[c]{High-Freq. \\ Routed Expert} \\
    \midrule
    % country\_capital\_city  
    capital & $\mathcal{H}^{22}_{13}$, $\mathcal{H}^{18}_{10}$, $\mathcal{H}^{23}_{15}$ & $\mathcal{E}^{20}_{59}$, $\mathcal{E}^{21}_{48}$, $\mathcal{E}^{21}_{14}$ \\ %(shared expert + top 0 routed expert)
    % country\_language  
    language & $\mathcal{H}^{22}_{13}$, $\mathcal{H}^{18}_{10}$, $\mathcal{H}^{23}_{15}$ & $\mathcal{E}^{20}_{59}$, $\mathcal{E}^{23}_{20}$, $\mathcal{E}^{21}_{20}$ \\ % (shared expert + only 2nd expert activated)
    % name\_birthplace
    birthplace
    & $\mathcal{H}^{22}_{13}$, $\mathcal{H}^{18}_{10}$, $\mathcal{H}^{19}_{12}$ & $\mathcal{E}^{20}_{59}$, $\mathcal{E}^{21}_{48}$, $\mathcal{E}^{18}_{38}$\\ % (w/o shared expert + top 1 routed expert)

    \bottomrule
  \end{tabular}}
  \caption{Top 3 attention heads and experts in Qwen 1.5-MoE.}
  \label{tab:attn_head_to_expert}
\end{table}

% \section{Knowledge-Expert Interaction Analysis}
\section{MoE Knowledge Localization: Trade-offs and Robustness}

In this section, we investigate the phenomenon of neuron-level knowledge localization in MoE models. 
% Our analysis focuses on the top 100 most important neurons, and the experts to whom they belong. 
For each knowledge type, identify the 100 most important neurons, determine which experts they belong to, and then we list the top 5 experts (i.e. the five experts containing the greatest number of those 100 neurons).
Here, our analysis is more concentrated on fine-grained experts, whose specialization trends are particularly pronounced \citep{wang-etal-2024-expert}; for coarse-grained MoE models such as Mixtral, we provide a detailed description in Appendix~\ref{app:complete_exp}. We provide insights into expert specialization (\mysection{sec:expert_specialization}) and investigate how architectural design affects the robustness and redundancy of the model, especially under expert failure scenarios (\mysection{sec:block_expert}). Through systematic expert-blocking experiments, we reveal critical trade-offs between specialization depth, task sensitivity, and fault tolerance.

\begin{table*}[t]
\centering
\resizebox{\textwidth}{!}{
\begin{tabular}{lllcccc}
\hline
\textbf{Relation} & \textbf{Model} & \textbf{Top5 Routed Experts} & \multicolumn{4}{c}{\textbf{MRR (Excl. Top$n$ Experts)}} \\ \cline{4-7}
 &  &  & \textbf{-} & \textbf{Top1} & \textbf{ Top5} & \textbf{Top10} \\ \hline

\multirow{2}{*}{name\_birthplace} & Qwen 1.5-MoE &  $\mathcal{E}^{20}_{59}$, $\mathcal{E}^{23}_{20}$, $\mathcal{E}^{19}_{42}$, $\mathcal{E}^{22}_{37}$, $\mathcal{E}^{16}_{13}$ & 0.85 & 0.84 & 0.83 & 0.81 \\ \cline{2-7}
                      & OLMoE &  $\mathcal{E}^{13}_{56}$, $\mathcal{E}^{15}_{49}$, $\mathcal{E}^{14}_{2}$, $\mathcal{E}^{12}_{40}$, $\mathcal{E}^{12}_{47}$ & 0.82 & 0.80 & 0.80 & 0.60 \\ \hline

\multirow{2}{*}{country\_capital\_city} & Qwen 1.5-MoE &$\mathcal{E}^{20}_{59}$, $\mathcal{E}^{21}_{48}$, $\mathcal{E}^{22}_{1}$, $\mathcal{E}^{21}_{14}$, $\mathcal{E}^{18}_{38}$ &  0.71 & 0.68 & 0.68 & 0.40 \\\cline{2-7}
                            & OLMoE & $\mathcal{E}^{13}_{56}$, $\mathcal{E}^{12}_{40}$, $\mathcal{E}^{14}_{2}$,  $\mathcal{E}^{13}_{52}$, $\mathcal{E}^{14}_{12}$ & 1.0 & 1.0 & 0.76 & 0.76 \\  \hline
\multirow{2}{*}{country\_language} & Qwen 1.5-MoE & $\mathcal{E}^{22}_{16}$, $\mathcal{E}^{20}_{59}$, $\mathcal{E}^{23}_{20}$, $\mathcal{E}^{21}_{20}$, $E^{19}_{3}$ & 0.94 & 0.92 & 0.92 & 0.92 \\ \cline{2-7}
                         & OLMoE &  $\mathcal{E}^{13}_{7}$, $\mathcal{E}^{14}_{3}$, $\mathcal{E}^{15}_{49}$, $\mathcal{E}^{12}_{0}$,  $\mathcal{E}^{13}_{56}$  & 0.96 & 0.92 & 0.87 & 0.68 \\ \hline
% \multirow{2}{*}{landmark\_in\_country} & Qwen 1.5-MoE &$E_{20}^{59}$, $E_{21}^{48}$, $E_{21}^{20}$, $E_{22}^{37}$, $E_{19}^{42}$& & & & \\ \cline{2-7}
%                          & OLMoE &  $E_{13}^{56}$, $E_{15}^{49}$, $E_{13}^{1}$, $E_{14}^{2}$, $E_{12}^{40}$ & & & & \\  \hline

\multirow{2}{*}{fruit\_inside\_color} & Qwen 1.5-MoE & $\mathcal{E}^{19}_{15}$, $\mathcal{E}^{22}_{38}$, $\mathcal{E}^{20}_{1}$, $\mathcal{E}^{22}_{15}$, $\mathcal{E}^{18}_{10}$ & 0.74 & 0.68 & 0.63 & 0.60 \\ \cline{2-7}
                            & OLMoE & $\mathcal{E}^{12}_{11}$, $\mathcal{E}^{14}_{8}$, $\mathcal{E}^{15}_{45}$, $\mathcal{E}^{12}_{32}$, $\mathcal{E}^{13}_{59}$ & 0.76 & 0.60 & 0.52 & 0.41 \\ \hline

\multirow{2}{*}{object\_superclass}  & Qwen 1.5-MoE & $\mathcal{E}^{22}_{11}$, $\mathcal{E}^{21}_{25}$, $\mathcal{E}^{21}_{25}$, $\mathcal{E}^{20}_{1}$, $\mathcal{E}^{19}_{15}$ & 0.83 & 0.82 & 0.80 & 0.79\\ \cline{2-7}
                            & OLMoE & $\mathcal{E}^{14}_{8}$, $\mathcal{E}^{13}_{59}$, $\mathcal{E}^{15}_{38}$, $\mathcal{E}^{14}_{53}$, $\mathcal{E}^{14}_{59}$ & 0.80 & 0.75 & 0.70 & 0.66 \\ \hline

\end{tabular}}
\caption{Top 5 experts %—among those that host the top 100 most important neurons—
for each knowledge type, along with the model’s MRR after excluding the top 1, top 5, and top 10 of those experts. The complete experimental results can be found in Appendix \ref{app:complete_exp}.}
\label{tab:excluding_expert}
\end{table*}

\subsection{Expert Specialization and Architectural Trade-offs} \label{sec:expert_specialization}

As shown in Table~\ref{tab:excluding_expert}, our findings reveal that certain experts consistently rank among the top five across various knowledge categories such as name\_birthplace, country\_capital\_city, and country\_language.

The specialization of experts in MoE models is profoundly influenced by architectural depth. In deeper architectures like Qwen 1.5-MoE, experts adopt a clear hierarchical division of labor. Take “\textit{The capital of Canada is the city of}” as an example. Early layers host generalists such as $\mathcal{E}^{20}_{59}$, which extract broad semantic features—for example, recognizing ``Canada'' as a geographic entity. Subsequent layers deploy specialists like $\mathcal{E}^{21}_{20}$, which refine these features into precise attributes (e.g., mapping ``Canada'' to ``Ottawa''). This stratification is evident in tasks such as geographic reasoning: when processing the input ``The capital of Canada is the city of'',
$\mathcal{E}^{20}_{59}$ first isolates the country context, while $\mathcal{E}^{21}_{20}$ generates the answer ``Ottawa'' through attribute association.

In contrast, shallow architectures like OLMoE face inherent constraints. With only 16 layers, experts such as 
$\mathcal{E}^{56}_{13}$must balance multiple roles, handling both geographic tasks (e.g., country-language mappings) and object attributes (e.g., fruit\_inside\_color). While this promotes cross-task generalization, it sacrifices fine-grained precision. For instance, 
$\mathcal{E}^{14}_{8}$ in OLMoE processes both fruit colors and object hierarchies, leading to blurred functional boundaries compared to Qwen 1.5-MoE’s specialized experts.

These architectural differences further impact robustness. Deep models leverage shared experts and layered redundancy to mitigate failures: blocking $\mathcal{E}^{21}_{20}$ in Qwen 1.5-MoE causes only a marginal MRR drop ($0.85 \to 0.84$) for geographic tasks. Shallow models, however, lack such safeguards. Blocking the critical expert 
$\mathcal{E}^{56}_{13}$ in OLMoE degrades MRR by 30\%, underscoring the vulnerability of architectures that prioritize breadth over depth. This trade-off highlights a fundamental design principle: deeper MoE models achieve robustness through hierarchical specialization, while shallower models rely on expert versatility at the cost of precision.

\subsection{Robustness and Redundancy in MoE Architectures}\label{sec:block_expert}
Table~\ref{tab:excluding_expert} shows the changes in MRR of Qwen 1.5-MoE and OLMoE models after blocking high-frequency experts under different knowledge types. Experimental results show that as the Top-1, Top-5, and Top-10 experts are blocked, Qwen 1.5-MoE and OLMoE show different robustness in different tasks.

The interplay between expert specialization and architectural depth profoundly impacts model robustness. For Qwen 1.5-MoE, its 24-layer design and shared expert integration enable distributed knowledge storage, buffering against expert failures. Blocking the top 10 experts in general tasks like name\_birthplace causes only a marginal MRR drop ($0.85\to0.81$), as redundant general experts (e.g., $\mathcal{E}^{20}_{59}$) compensate for losses. However, geographic tasks like country\_capital\_city exhibit fragility: blocking core experts (e.g., $\mathcal{E}^{21}_{48}$) slashes MRR by 43\% (0.71 → 0.40), highlighting their reliance on specialized modules.

In contrast, OLMoE’s shallow 16-layer architecture lacks such safeguards. While its routing mechanism dynamically selects alternative experts to maintain performance when blocking top 1–5 experts (MRR remains 1.0 for country\_capital\_city), deeper dependencies overwhelm its redundancy capacity. Blocking the top 10 experts degrades MRR to 0.76, with geographic tasks like country\_language suffering a 30\% decline (0.96 → 0.68). This stark contrast underscores that depth enables hierarchical redundancy, whereas shallow models trade transient adaptability for long-term fragility.

Furthermore, task sensitivity further dictates robustness requirements. \textbf{Core-sensitive tasks} (e.g., geography) demand concentrated expertise. Qwen 1.5-MoE’s late-layer specialists (e.g., $\mathcal{E}^{21}_{48}$) are irreplaceable, necessitating redundancy through shared experts.
\textbf{Distributed-tolerant tasks} (e.g., object attributes) leverage broader expert participation. Qwen 1.5-MoE’s MRR drops mildly (0.83 → 0.79) when blocking experts for object\_superclass, while OLMoE plummets (0.80 → 0.66) due to overburdened generalists.

These findings inform actionable design principles: (1) Deep MoE architectures should prioritize shared experts in early layers for cross-task redundancy and allocate specialists to late layers for core-sensitive tasks.
(2) Shallow MoE models must balance expert versatility and routing adaptability, though their limited depth inherently restricts robustness.
(3) By aligning architectural depth with task requirements, MoE models can optimize both performance and reliability—a strategy exemplified by Qwen 1.5-MoE’s resilience and OLMoE’s adaptability trade-offs.

\section{Related Work}

\paragraph{Mixture of Experts}
The Mixture of Experts (MoE) framework, introduced by \citet{jacobs1991adaptive}, dynamically activates specialized sub-networks for inputs. Early applications in language modeling used LSTMs \citep{shazeer2017outrageously}, but recent work integrates MoE into Transformer FFN layers \citep{vaswani2017attention}, scaling parameters via sparse routing (e.g., top-$k$ gating \citep{fedus-2021-switch, lepikhin2021gshard,  roller2021hash,dai-etal-2022-stablemoe, zeng-etal-2024-turn, xue2024openmoe}). Innovations like DeepSeekMoE \citep{dai-etal-2024-deepseekmoe} and Qwen-MoE \citep{qwen_moe} enhance specialization through shared experts, balancing efficiency and performance.

\paragraph{Knowledge Attribution}
Studies on how models store knowledge reveal that Transformer MLPs act as key-value memories \citep{geva-etal-2021-transformer, dai-etal-2022-knowledge, meng2022locating, wu-etal-2023-depn, xu-etal-2023-language-representation, meng2023masseditingmemorytransformer}, with knowledge accumulating across layers \citep{geva-etal-2022-transformer, lv2024interpreting}. While methods like knowledge circuits \citep{yao2024knowledge} and neuron-level attribution \citep{yu-ananiadou-2024-neuron} interpret dense models, sparse MoE architectures—with dynamic routing and expert collaboration—remain underexplored. Our work bridges this gap, offering the first comprehensive analysis of MoE interpretability.

\section{Conclusion}
% This paper proposes a cross-level knowledge attribution algorithm for MoE models, filling the gap in existing methods (mainly for dense models). By comparing and analyzing MoE and dense models, the unique patterns of MoE (such as "mid-term start-up, late amplification" and "base-optimization" collaborative framework) are revealed, providing a new perspective for model interpretability. For the first time, the knowledge localization mechanism of heterogeneous MoE architecture (including shared experts) is systematically explored, and the semantically driven collaborative relationship between attention heads and experts is verified. The research results provide a theoretical basis for optimizing the MoE architecture (such as redundant design and expert division of labor), which is of practical value for improving model robustness and scenario adaptability.

In this paper, we propose a cross-level knowledge attribution algorithm for MoE models, bridging the interpretability gap between sparse and dense architectures. Through comparative analysis, we uncover MoE's unique efficiency patterns—such as “mid-stage activation, late-stage amplification” and a “basic-refinement” collaboration framework—where shared experts handle general tasks while routed experts specialize in refinement. We further validate a semantic-driven routing mechanism: attention heads coordinate expert selection via high temporal correlations, enabling task-aware processing.

Our findings offer actionable guidelines for MoE design:
\begin{itemize}
    \item Prioritize shared experts in early layers to provide cross-task redundancy and foundational processing.
    \item Allocate routed experts to deeper layers for domain-specific refinement, balancing specialization and efficiency.
    \item Optimize routing depth based on task sensitivity: Deploy deeper architectures for core-sensitive tasks (e.g., geography) and shallower models for distributed-tolerant tasks (e.g., object attributes).
    \item Enhance semantic alignment between attention heads and experts to improve routing robustness.
\end{itemize}

These principles address critical challenges in MoE scalability and reliability, offering a pathway to deploy efficient, interpretable models in resource-constrained scenarios. % Future work will extend our attribution algorithm to dynamic routing architectures and larger-scale models.

\section*{Acknowledgments}
This work was supported by the CAAI-Ant Group Research Fund; Guangdong Provincial Department of Education Project (Grant No.2024KQNCX028);  Scientific Research Projects for the Higher-educational Institutions (Grant No.2024312096), Education Bureau of Guangzhou Municipality; Guangzhou-HKUST(GZ) Joint Funding Program (Grant No.2025A03J3957), Education Bureau of Guangzhou Municipality.

\section*{Limitations}

Our analysis is confined to 7 B-parameter, static-routing MoE models, so it remains to be seen whether much larger (e.g., 100 B+) \citep{deepseekai2024deepseekv3technicalreport, sun2024hunyuanlargeopensourcemoemodel} or fully dynamic MoE \citep{huang-etal-2024-harder, guo2024dynamic} architectures will exhibit the same specialization patterns. We also omit retrieval-augmented MoEs (e.g., Monet \citep{park2025monet}), whose external-knowledge coupling adds confounding factors beyond our scope.

% \bibliography{custom}
\bibliography{acl_latex}
\newpage
\newpage
\appendix

\begin{table*}[t]
\centering
\resizebox{\textwidth}{!}{
\begin{tabular}{lcccccc}
\hline
\toprule
\makecell[l]{Model Name} & \makecell[c]{Layers} & \makecell[c]{FFN / Expert \\ Dimension} & \makecell[c]{Number of \\ Experts} & \makecell[c]{Routing \\ Strategy} & \makecell[c]{Shared \\ Expert} & \makecell[c]{Shared Expert \\ Dimension} \\ \midrule
LLama-7B    & 32     & 11008                        & -              & -             & -            & -                   \\
Qwen 1.5-7B     & 32     & 11008                        & -              & -             & -            & -                   \\
Mistral-7B & 32 & 14336 & - & - & - & - \\
Qwen1.5-MoE-A2.7B    & 24     & 1408                    & 60                & Top-4            & Yes          & 5632                   \\
OLMoE-1B-7B           & 16     & 1024                     & 64                & Top-8            & -            & -                   \\ 
Mixtral-8x7B  & 32 & 14336 & 8 & top-2 & - & - \\
\bottomrule
\end{tabular}}
\caption{Model Configuration.}
\label{tab:model_config}
\end{table*}

\section{Experiment Settings}
\label{sec:appendix_experiment}

% \subsection{Experiments}
\paragraph{Models}
We evaluate multiple models to ensure robust conclusions. These include three dense models (Llama-7B \citep{touvron2023llama1}, Qwen 1.5-7B \citep{qwen2023dense}) and Mistral-7B \citep{jiang2023mistral7b} and three Mixture-of-Experts (MoE) models (Qwen 1.5-MoE \citep{qwen_moe}, OLMoE \citep{muennighoff2024olmoe} and Mixtral 8x7B \citep{albert-2024-mixtral}). Qwen 1.5-MoE is a 24-layer model where each MoE layer comprises a shared expert and 60 routing experts, employing a Top-4 routing strategy during inference. Mixtral has 32 layers, 8 experts per layer, no shared experts, employing a Top-2 routing strategy. In contrast, OLMoE is a shallower model with 16 layers, where each MoE layer contains 64 routing experts and employs a Top-8 routing mechanism. These structural differences among the MoE models provide complementary perspectives, enriching our analysis.
Table~\ref{tab:model_config} lists the specific configuration comparison of each model.

\paragraph{Datasets}
We use the dataset provided by LRE \citep{hernandezlinearity}, which contains a diverse set of knowledge types, including linguistic patterns, common sense, factual information, and biases. A total of 12 distinct knowledge categories are selected for our experiments from the dataset (i.e., \textit{adjective\_antonym, word\_first\_letter, word\_last\_letter relations, object\_superclass, fruit\_inside\_color, work\_location relations, country\_language, country\_capital, name\_birthplace, name\_religion, occupation\_age,} and \textit{occupation\_gender}). This dataset serves two purposes: it provides explicit knowledge for the model and is also used to evaluate its zero-shot performance across varied knowledge domains.

\paragraph{Metrics}
We assess model performance using two standard metrics: HIT@10 and Mean Reciprocal Rank (MRR). HIT@10 measures the proportion of test instances for which the correct answer is ranked among the top 10 predictions:
\begin{equation}
    \text{HIT}@10 = \frac{1}{|V|} \sum_{i=1}^{|V|} \mathbb{I}(\text{rank}_i \leq 10),
\end{equation}
where \( \mathbb{I}(\cdot) \) is the indicator function, returning 1 if the condition is satisfied and 0 otherwise, and \( \text{rank}_i \) denotes the rank position of the correct answer for the \( i \)-th instance.

MRR computes the average of the reciprocal ranks of the correct answers:
\begin{equation}
    \text{MRR} = \frac{1}{|V|} \sum_{i=1}^{|V|} \frac{1}{\text{rank}_i}.
\end{equation}

While HIT@10 emphasizes the model’s ability to retrieve correct answers within the top predictions, MRR provides a more sensitive measure by giving higher scores to cases where the correct answer is ranked highly. Together, these metrics offer complementary insights into the overall ranking performance of the model.

\begin{figure}[t]
  \includegraphics[width=\columnwidth]{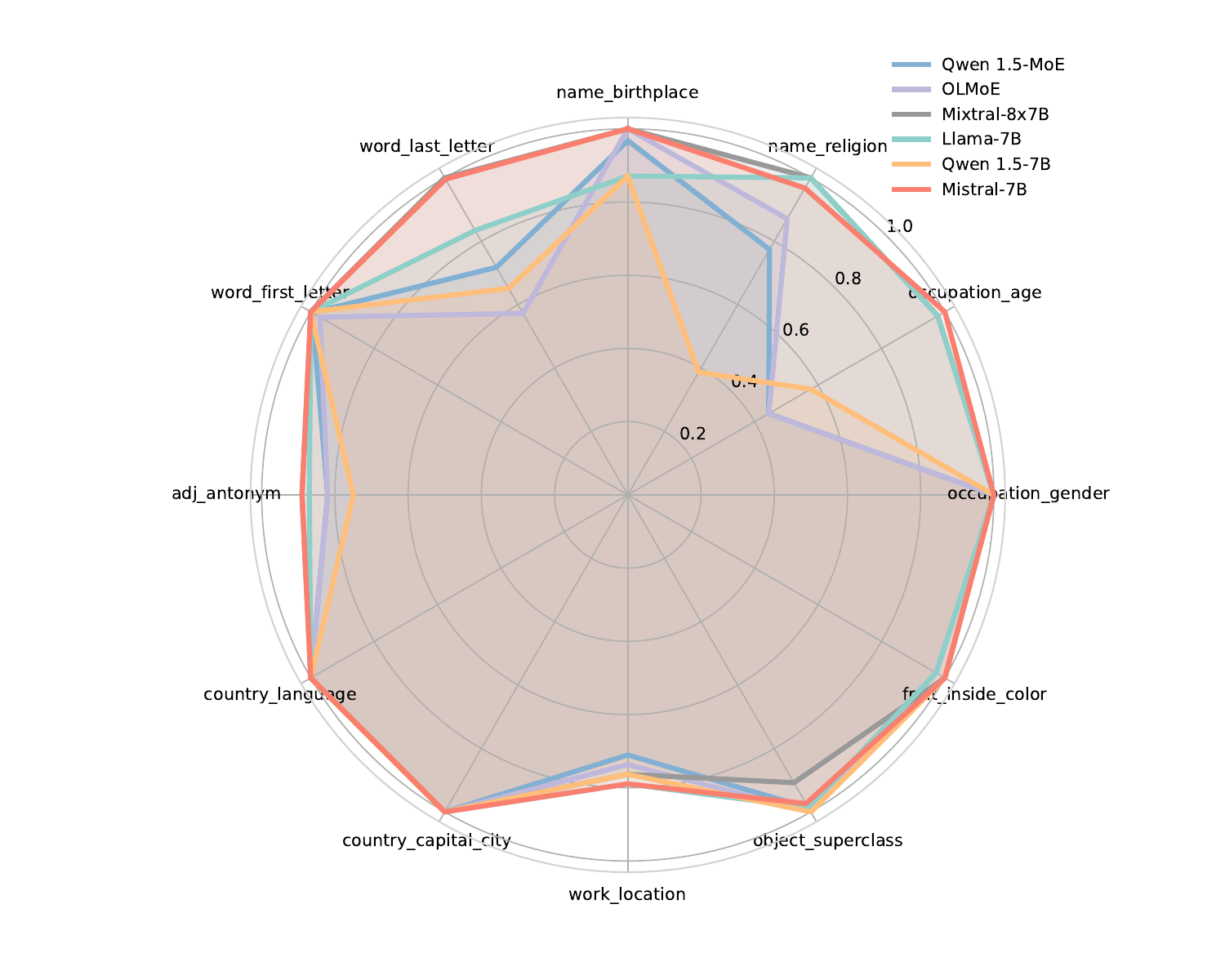}
  \caption{HIT@10 scores of four models on 12 different relations.}
  \label{fig:overall_experiments_result}
\end{figure}

\paragraph{Results}
Figure~\ref{fig:overall_experiments_result} shows the HIT@10 of different models on 12 tasks. It can be seen that the performance of different models varies greatly in some types. Therefore, when analyzing the experimental results, we mainly focus on the relations with similar performance to avoid the influence of the model's internal parameter knowledge on the effect of knowledge attribution.

\begin{figure}[t]
  \includegraphics[width=\columnwidth]{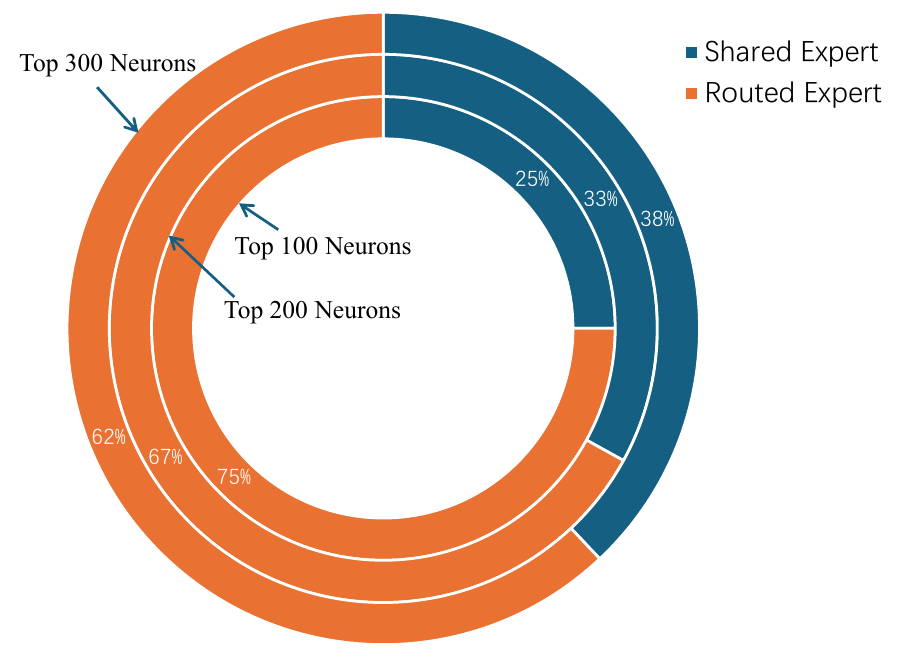}
  \caption{Type distribution of neurons in the top 100, 200, and 300 of Qwen 1.5-MoE importance scores.}
  \label{fig:pie}
\end{figure}

\section{Detailed Analysis of Coarse/Fine-Grained Routed Experts}
\label{app:complete_exp}
\begin{figure}[t]
  \includegraphics[width=\columnwidth]{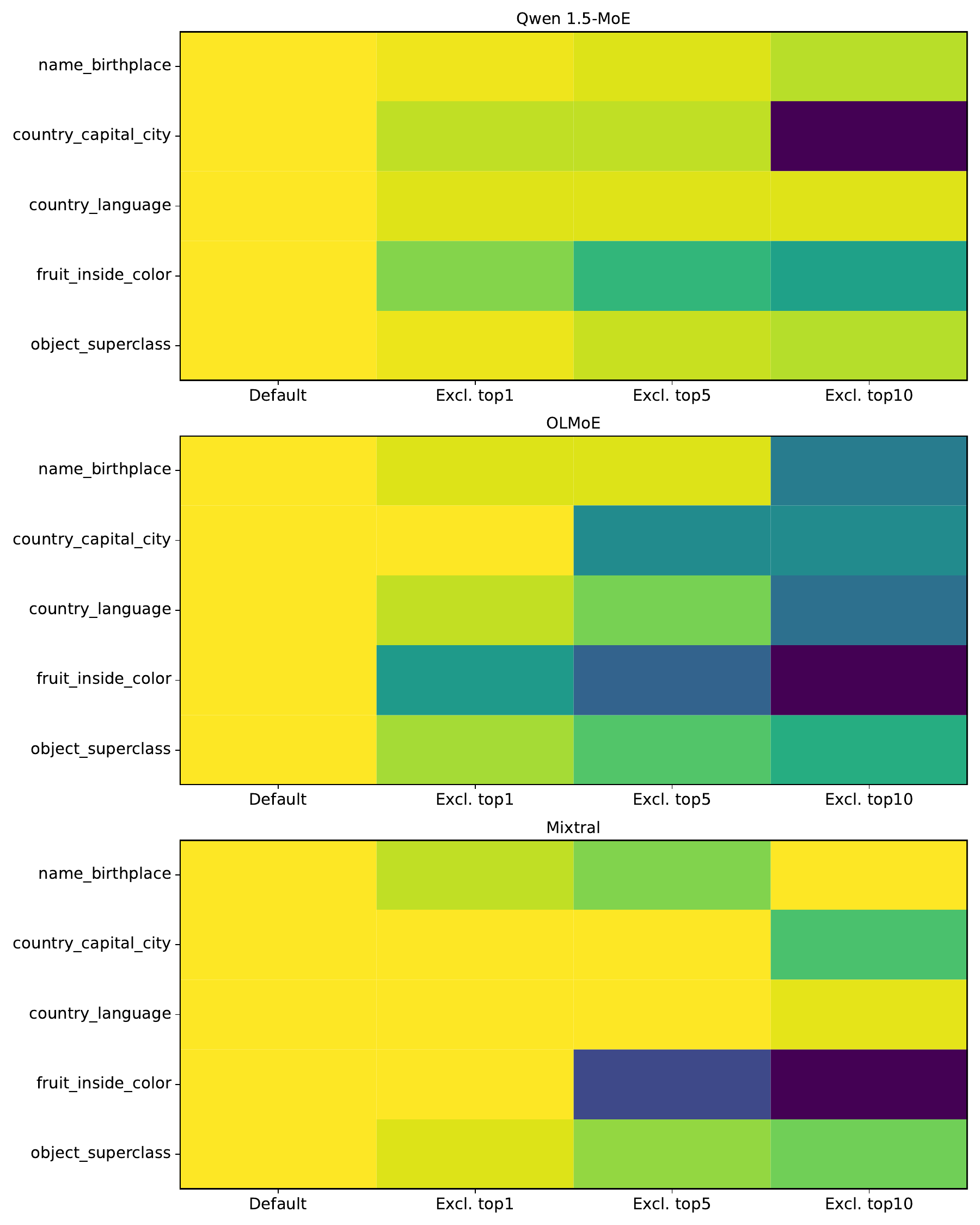}
  \caption{Robustness comparison heat map. The darker the color, the greater the drop in MRR.}
  \label{fig:moe_heatmap}
\end{figure}

We report the full expert-isolation results in Table~\ref{tab:summary}. Figure~\ref{fig:moe_heatmap} visualizes the heat map of the MRR drop of models under different settings. As expected, the two fine-grained MoE models suffer precipitous drops in performance when their top specialists are blocked—reaffirming the strong localization of functionality we described in \mysection{sec:block_expert}. By contrast, Mixtral—a coarse-grained MoE with only eight experts per layer and a top-2 routing policy—maintains near-identical accuracy even when its ten most “important” experts are silenced.

This resilience stems from Mixtral’s relatively dense routing: with two experts activated per token, each expert has a $1/4$ chance of selection, compared to just $1/15$ in Qwen 1.5-MoE and $1/8$ in OLMoE. Over long sequences, this routing density means that almost every expert participates, distributing representational capacity broadly rather than concentrating it in a few specialists. 
Since the Mixtral authors do not explicitly detail their expert initialization, we tentatively infer—based on related research \citep{lo-etal-2025-closer}—that its experts may be seeded by copying FFN weights from the dense Mistral model, a practice that would likely homogenize their behaviors and impede the development of fine-grained specialization.

Taken together, these findings highlight a fundamental trade-off in MoE design. Coarse-grained architectures like Mixtral effectively amplify model capacity and fault tolerance by maximizing activation coverage, but they do so at the expense of individual expert specialization. Fine-grained MoEs, on the other hand, cultivate highly specialized experts—yielding sharp performance gains on focused tasks—but are more vulnerable to expert failure. Future work might explore hybrid schemes that combine a base level of broad activation with targeted specialization, for instance by increasing expert count, adjusting routing sparsity, or employing specialized initialization and regularization to encourage functional diversity among experts.

\begin{table}[t]
\centering
\resizebox{\columnwidth}{!}{
\begin{tabular}{l c} % 让第二列居中
\hline
\textbf{Category} & \multicolumn{1}{c}{\textbf{Top-5 Routed Shared Expert Neurons}} \\ \hline
birthplace & $\mathcal{N}^{20}_{1044}$, $\mathcal{N}^{16}_{204}$, $\mathcal{N}^{21}_{711}$, $\mathcal{N}^{17}_{123}$, $\mathcal{N}^{19}_{641}$ \\ \hline
capital & $\mathcal{N}^{20}_{5}$, $\mathcal{N}^{15}_{1270}$, $\mathcal{N}^{15}_{415}$, $\mathcal{N}^{9}_{1060}$, $\mathcal{N}^{18}_{637}$ \\ \hline
language & $\mathcal{N}^{18}_{1137}$, $\mathcal{N}^{18}_{232}$, $\mathcal{N}^{17}_{682}$, $\mathcal{N}^{15}_{647}$, $\mathcal{N}^{19}_{322}$ \\ \hline
fruit & $\mathcal{N}^{20}_{947}$, $\mathcal{N}^{16}_{413}$, $\mathcal{N}^{15}_{173}$, $\mathcal{N}^{19}_{1158}$, $\mathcal{N}^{19}_{900}$ \\ \hline
object & $\mathcal{N}^{17}_{202}$, $\mathcal{N}^{18}_{1145}$, $\mathcal{N}^{20}_{631}$, $\mathcal{N}^{21}_{1262}$, $\mathcal{N}^{20}_{359}$ \\ \hline
\end{tabular}}
\caption{Top-5 shared expert neurons for each relation. For example, $\mathcal{N}^{20}_{1044}$ represents the 1044th neuron of the shared expert in the 20th layer.}
\label{tab:shared_neuron}
\end{table}

\begin{table*}[t]
\centering
\resizebox{\textwidth}{!}{
\begin{tabular}{lllcccc}
\hline
\textbf{Relation} & \textbf{Model} & \textbf{Top5 Routed Experts} & \multicolumn{4}{c}{\textbf{MRR (Excl. Top$n$ Experts)}} \\ \cline{4-7}
 &  &  & \textbf{-} & \textbf{Top1} & \textbf{Top5} & \textbf{Top10} \\ \hline
 
\multirow{3}{*}{country\_capital\_city} & Qwen 1.5-MoE &$\mathcal{E}^{20}_{59}$, $\mathcal{E}^{21}_{48}$, $\mathcal{E}^{22}_{1}$, $\mathcal{E}^{21}_{14}$, $\mathcal{E}^{18}_{38}$ &  0.71 & 0.68 & 0.68 & 0.40 \\\cline{2-7}
                            & OLMoE & $\mathcal{E}^{13}_{56}$, $\mathcal{E}^{12}_{40}$, $\mathcal{E}^{14}_{2}$,  $\mathcal{E}^{13}_{52}$, $\mathcal{E}^{14}_{12}$ & 1.0 & 1.0 & 0.76 & 0.76 \\ \cline{2-7}
                            & Mixtral & $\mathcal{E}^{24}_{7}$, $\mathcal{E}^{25}_{0}$, $\mathcal{E}^{22}_{1}$, $\mathcal{E}^{26}_{0}$, $\mathcal{E}^{19}_{2}$ & 1.0 & 1.0 & 1.0 & 0.93 \\ \hline
                            
\multirow{3}{*}{country\_language} & Qwen 1.5-MoE & $\mathcal{E}^{22}_{16}$, $\mathcal{E}^{20}_{59}$, $\mathcal{E}^{23}_{20}$, $\mathcal{E}^{21}_{20}$, $E^{19}_{3}$ & 0.94 & 0.92 & 0.92 & 0.92 \\ \cline{2-7}
                         & OLMoE &  $\mathcal{E}^{13}_{7}$, $\mathcal{E}^{14}_{3}$, $\mathcal{E}^{15}_{49}$, $\mathcal{E}^{12}_{0}$,  $\mathcal{E}^{13}_{56}$  & 0.96 & 0.92 & 0.87 & 0.68 \\ \cline{2-7}
                         & Mixtral & $\mathcal{E}^{22}_{5}$, $\mathcal{E}^{24}_{4}$, $\mathcal{E}^{19}_{6}$, $\mathcal{E}^{25}_{7}$, $\mathcal{E}^{21}_{0}$ & 0.98 & 0.98 & 1.0 & 0.97 \\ \hline       
                         
\multirow{3}{*}{adj\_antonym}  & Qwen 1.5-MoE & $\mathcal{E}^{20}_{27}$, $\mathcal{E}^{22}_{15}$, $\mathcal{E}^{22}_{59}$, $\mathcal{E}^{16}_{49}$, $\mathcal{E}^{21}_{4}$ & 0.60 & 0.57 & 0.50 & 0.53 \\ \cline{2-7}
                            & OLMoE & $\mathcal{E}^{11}_{24}$, $\mathcal{E}^{13}_{49}$, $\mathcal{E}^{12}_{32}$, $\mathcal{E}^{12}_{36}$, $\mathcal{E}^{14}_{25}$ & 0.67 & 0.64 & 0.65 & 0.58 \\ \cline{2-7}
                            & Mixtral & $\mathcal{E}^{20}_{0}$, $\mathcal{E}^{19}_{6}$, $\mathcal{E}^{23}_{5}$, $\mathcal{E}^{31}_{6}$, $\mathcal{E}^{21}_{5}$ & 0.75 & 0.75 & 0.74 & 0.73 \\ \hline                            
\multirow{3}{*}{word\_first\_letter}  & Qwen 1.5-MoE & $\mathcal{E}^{19}_{7}$, $\mathcal{E}^{16}_{41}$, $\mathcal{E}^{16}_{49}$, $\mathcal{E}^{18}_{41}$, $\mathcal{E}^{19}_{45}$ & 0.96 & 0.95 & 0.93 & 0.92\\ \cline{2-7}
                            & OLMoE & $\mathcal{E}^{14}_{39}$, $\mathcal{E}^{15}_{17}$, $\mathcal{E}^{12}_{45}$, $\mathcal{E}^{12}_{32}$, $\mathcal{E}^{15}_{60}$ & 0.79 & 0.75 & 0.68 & 0.48 \\ \cline{2-7}
                            & Mixtral & $\mathcal{E}^{30}_{7}$, $\mathcal{E}^{29}_{7}$, $\mathcal{E}^{28}_{0}$, $\mathcal{E}^{27}_{4}$, $\mathcal{E}^{28}_{3}$ & 1.0 & 1.0 & 1.0 & 0.99\\ \hline
                            
\multirow{3}{*}{word\_last\_letter}  & Qwen 1.5-MoE & $\mathcal{E}^{19}_{7}$, $\mathcal{E}^{16}_{41}$, $\mathcal{E}^{16}_{49}$, $\mathcal{E}^{19}_{45}$, $\mathcal{E}^{18}_{41}$ & 0.27 & 0.25 & 0.12 & 0.12 \\ \cline{2-7}
                            & OLMoE & $\mathcal{E}^{14}_{7}$, $\mathcal{E}^{12}_{32}$, $\mathcal{E}^{15}_{17}$, $\mathcal{E}^{11}_{7}$, $\mathcal{E}^{10}_{62}$ & 0.12 & 0.12 & 0.10 & 0.08 \\ \cline{2-7}
                            & Mixtral & $\mathcal{E}^{30}_{7}$, $\mathcal{E}^{29}_{7}$, $\mathcal{E}^{28}_{0}$, $\mathcal{E}^{27}_{4}$, $\mathcal{E}^{24}_{7}$ & 0.95 & 0.95 & 0.91 & 0.68 \\ \hline      
                            
\multirow{3}{*}{name\_birthplace} & Qwen 1.5-MoE &  $\mathcal{E}^{20}_{59}$, $\mathcal{E}^{23}_{20}$, $\mathcal{E}^{19}_{42}$, $\mathcal{E}^{22}_{37}$, $\mathcal{E}^{16}_{13}$ & 0.85 & 0.84 & 0.83 & 0.81 \\ \cline{2-7}
                      & OLMoE &  $\mathcal{E}^{13}_{56}$, $\mathcal{E}^{15}_{49}$, $\mathcal{E}^{14}_{2}$, $\mathcal{E}^{12}_{40}$, $\mathcal{E}^{12}_{47}$ & 0.82 & 0.80 & 0.80 & 0.60 \\ \cline{2-7}
                      & Mixtral &  $\mathcal{E}^{18}_{1}$, $\mathcal{E}^{22}_{1}$, $\mathcal{E}^{25}_{3}$, $\mathcal{E}^{29}_{4}$, $\mathcal{E}^{31}_{2}$ & 0.87 & 0.85 & 0.83 & 0.87\\ \hline 
                      
\multirow{3}{*}{name\_religion} & Qwen 1.5-MoE &  $\mathcal{E}^{20}_{30}$, $\mathcal{E}^{23}_{20}$, $\mathcal{E}^{16}_{13}$, $\mathcal{E}^{20}_{53}$, $\mathcal{E}^{18}_{38}$ & 0.36 & 0.45 & 0.32 & 0.37 \\ \cline{2-7}
                      & OLMoE &  $\mathcal{E}^{13}_{14}$, $\mathcal{E}^{12}_{37}$, $\mathcal{E}^{12}_{47}$, $\mathcal{E}^{13}_{1}$, $\mathcal{E}^{15}_{49}$ & 0.44 & 0.46 & 0.45 & 0.44 \\ \cline{2-7}
                      & Mixtral & $\mathcal{E}^{19}_{4}$, $\mathcal{E}^{20}_{5}$, $\mathcal{E}^{22}_{5}$, $\mathcal{E}^{24}_{7}$, $\mathcal{E}^{18}_{1}$ & 0.87 & 0.86 & 0.84 & 0.82 \\ \hline
                      
\multirow{3}{*}{occupation\_age} & Qwen 1.5-MoE &$\mathcal{E}^{20}_{46}$, $\mathcal{E}^{19}_{48}$, $\mathcal{E}^{21}_{3}$, $\mathcal{E}^{18}_{10}$, $\mathcal{E}^{22}_{9}$ &  0.22 & 0.21 & 0.36 & 0.36\\\cline{2-7}
                            & OLMoE & $\mathcal{E}^{12}_{27}$, $\mathcal{E}^{13}_{26}$, $\mathcal{E}^{14}_{56}$,  $\mathcal{E}^{15}_{62}$, $\mathcal{E}^{11}_{53}$ & 0.36 & 0.40 & 0.37 & 0.40 \\  \cline{2-7}
                            & Mixtral & $\mathcal{E}^{29}_{0}$, $\mathcal{E}^{30}_{4}$, $\mathcal{E}^{20}_{0}$, $\mathcal{E}^{31}_{6}$, $\mathcal{E}^{31}_{4}$ & 0.35 & 0.37 & 0.49 & 0.58\\ \hline
                            
\multirow{3}{*}{occupation\_gender} & Qwen 1.5-MoE & $\mathcal{E}^{22}_{42}$, $\mathcal{E}^{17}_{4}$, $\mathcal{E}^{19}_{29}$, $\mathcal{E}^{20}_{46}$, $\mathcal{E}^{18}_{10}$ & 0.95 & 0.96 & 0.94 & 0.96 \\ \cline{2-7}
                         & OLMoE &  $\mathcal{E}^{12}_{27}$, $\mathcal{E}^{13}_{44}$, $\mathcal{E}^{13}_{1}$, $\mathcal{E}^{13}_{26}$,  $\mathcal{E}^{15}_{14}$  & 0.95 & 0.86 & 0.84 & 0.55 \\ \cline{2-7}
                         & Mixtral & $\mathcal{E}^{23}_{0}$, $\mathcal{E}^{20}_{0}$, $\mathcal{E}^{19}_{6}$, $\mathcal{E}^{25}_{6}$, $\mathcal{E}^{29}_{0}$ & 0.97 & 1.0 & 0.97 & 0.97\\ \hline
                         
\multirow{3}{*}{fruit\_inside\_color} & Qwen 1.5-MoE & $\mathcal{E}^{19}_{15}$, $\mathcal{E}^{22}_{38}$, $\mathcal{E}^{20}_{1}$, $\mathcal{E}^{22}_{15}$, $\mathcal{E}^{18}_{10}$ & 0.74 & 0.68 & 0.63 & 0.60 \\ \cline{2-7}
                            & OLMoE & $\mathcal{E}^{12}_{11}$, $\mathcal{E}^{14}_{8}$, $\mathcal{E}^{15}_{45}$, $\mathcal{E}^{12}_{32}$, $\mathcal{E}^{13}_{59}$ & 0.76 & 0.60 & 0.52 & 0.41 \\ \cline{2-7}
                            & Mixtral & $\mathcal{E}^{27}_{4}$, $\mathcal{E}^{26}_{2}$, $\mathcal{E}^{24}_{3}$, $\mathcal{E}^{23}_{5}$, $\mathcal{E}^{29}_{0}$ & 0.74 & 0.75 & 0.60 & 0.56 \\ \hline
                            
\multirow{3}{*}{object\_superclass}  & Qwen 1.5-MoE & $\mathcal{E}^{22}_{11}$, $\mathcal{E}^{21}_{25}$, $\mathcal{E}^{21}_{25}$, $\mathcal{E}^{20}_{1}$, $\mathcal{E}^{19}_{15}$ & 0.83 & 0.82 & 0.80 & 0.79\\ \cline{2-7}
                            & OLMoE & $\mathcal{E}^{14}_{8}$, $\mathcal{E}^{13}_{59}$, $\mathcal{E}^{15}_{38}$, $\mathcal{E}^{14}_{53}$, $\mathcal{E}^{14}_{59}$ & 0.80 & 0.75 & 0.70 & 0.66 \\ \cline{2-7}
                            & Mixtral & $\mathcal{E}^{20}_{6}$, $\mathcal{E}^{31}_{3}$, $\mathcal{E}^{21}_{2}$, $\mathcal{E}^{19}_{6}$, $\mathcal{E}^{24}_{4}$ & 0.76 & 0.75 & 0.73 & 0.72\\ \hline

\multirow{3}{*}{work\_location}  & Qwen 1.5-MoE & $\mathcal{E}^{18}_{51}$, $\mathcal{E}^{22}_{8}$, $\mathcal{E}^{17}_{47}$, $\mathcal{E}^{19}_{27}$, $\mathcal{E}^{22}_{54}$ & 0.45 & 0.47 & 0.51 & 0.51\\ \cline{2-7}
                            & OLMoE & $\mathcal{E}^{14}_{43}$, $\mathcal{E}^{12}_{0}$, $\mathcal{E}^{13}_{63}$, $\mathcal{E}^{13}_{13}$, $\mathcal{E}^{12}_{25}$ & 0.52 & 0.48 & 0.45 & 0.52 \\ \cline{2-7}
                            & Mixtral & $\mathcal{E}^{21}_{1}$, $\mathcal{E}^{20}_{5}$, $\mathcal{E}^{23}_{0}$, $\mathcal{E}^{24}_{0}$, $\mathcal{E}^{28}_{1}$ & 0.61 & 0.61 & 0.59 & 0.58 \\ \hline

\end{tabular}}
\caption{Top-5 most frequently selected experts for each knowledge type, along with the MRR of the model after excluding the top 1, 5, and 10 experts. }
\label{tab:summary}
\end{table*}

\section{Cross-Knowledge Neuron Dynamics}

In the context of shared experts, we observe a significant overlap of neurons for different samples within the same relation (Table~\ref{tab:shared_neuron}). For instance, in the "capital" task, neurons responsible for predicting the capitals of various countries often share common activations. This suggests that neurons can store knowledge specific to a given relation and share it across different instances. However, the overlap of neurons across different relations remains minimal. For example, the shared neuron overlap between the "capital" and "language" tasks is typically less than 5\%. This indicates that the role of shared experts is task-specific, with distinct mechanisms for knowledge storage and processing across tasks.

Further analysis, as shown in Table~\ref{tab:summary}, reveals that the effect significantly increases after blocking the top 10 experts for certain relations (e.g., the MRR for the occupation\_age relation in Qwen 1.5-MoE changes from 0.22 to 0.36). This counterintuitive phenomenon contrasts with the behavior observed in other relations, where blocking commonly used experts leads to a decrease in performance or fluctuations around baseline values. However, it is important to note that the model's baseline MRR on this particular task is quite low and close to random, meaning that the observed increase does not indicate an issue with our method.

Moreover, when analyzing the top 100 neurons, we find that approximately 25\% of these neurons belong to routed experts, while the remaining 75\% are associated with shared experts, as shown in Figure~\ref{fig:pie}. However, as we expand the analysis to the top 300 neurons, the proportion of routed experts increases to 38\%. This suggests that more expert collaboration is required for handling complex tasks, with the routing mechanism dynamically expanding its activation range. As task complexity increases, the routing mechanism activates additional experts, creating a dynamic collaboration chain that connects core experts with auxiliary experts. This mechanism enables the model to flexibly coordinate knowledge from a broader set of experts, ultimately improving its performance on more complex tasks.

\end{document}